1# Artificial Color Constancy via GoogLeNet with Angular Loss Function

Oleksii Sidorov
The Norwegian Colour and Visual Computing Laboratory
NTNU
Gjovik, Norway
oleksiis@stud.ntnu.no
**Abstract**

Color Constancy is the ability of the human visual system to perceive colors unchanged independently of the illumination. Giving a machine this feature will be beneficial in many fields where chromatic information is used. Particularly, it significantly improves scene understanding and object recognition.

In this paper, we propose transfer learning-based algorithm, which has two main features: accuracy higher than many state-of-the-art algorithms and simplicity of implementation. Despite the fact that GoogLeNet was used in the experiments, given approach may be applied to any CNN. Additionally, we discuss design of a new loss function oriented specifically to this problem, and propose a few the most suitable options.
## 1 Introduction

Color is an important part of visual information. However, color is not an intrinsic feature of an object, but the result of interaction between scene illumination, object's reflection, camera sensor's sensitivity, etc. Since most applications require only the object's intrinsic characteristics, separation of this information (particularly, removing illumination color casts) is an important task.

Human visual system solves this task via Color Constancy (CC) – complex mechanism which involves color adaptation, color memory, and other features of human vision. Creation of artificial algorithm which is able to do the same would be beneficial for many computer vision applications. Scene understanding, object recognition, pattern recognition, stereo vision, tracking, quality control, and many other fields use chromatic information, and may suffer from its falseness. For example, Hosseini and Poovendran [26] have illustrated how VGG-16 network [36] can be "fooled" by wrong colors (Fig. 1).

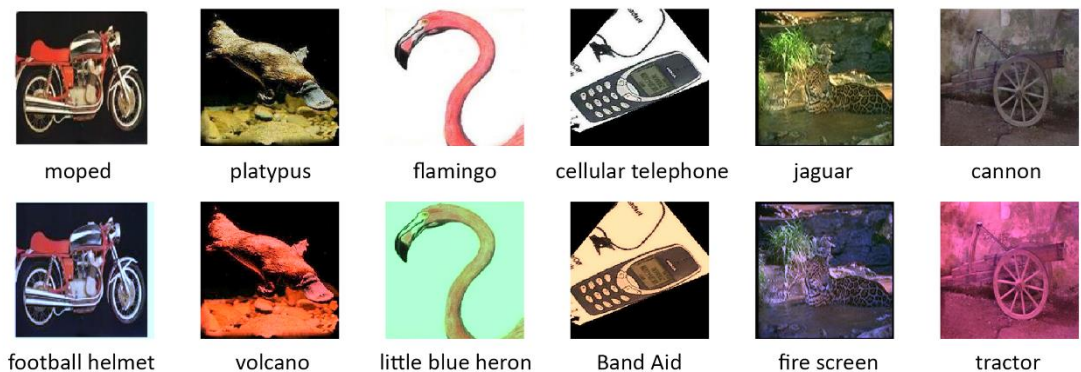

**Figure 1**. Classification of images by VGG-16 net. Top row: original images from Caltech 101 dataset [14]; bottom row: the same images casted by random uniform illumination.

Despite the high importance of this problem, universal solution still has not been found. Recently, development of Machine Learning techniques and especially Convolutional Neural Networks (CNNs) facilitated creation of more accurate CC-algorithms. Considering that majority of original CNNs designed specifically for Color Constancy are quite simple and consist of only a few layers, we propose to improve their efficiency by means of more deep and more powerful nets using transfer learning approach which is widely used in deep learning. Besides of gain in complexity, this approach allows to greatly reduce training time, and gives an opportunity to people which are not familiar with deep learning and cannot design CNN from scratch use efficient algorithms for their needs. (The shortest form of our algorithm is only 30 lines of code long and takes just a few hours to train).

The general framework of CNN-based Color Constancy methods is an image regression which predicts coordinates of an illumination vector. Since a length of a vector is normalized in the result only its orientation is important. Consequently, we propose instead of Mean Squared Error (MSE) which tries to fit both orientation and length, use as loss function angular loss which considers only orientation of a vector, and thus adds flexibility in parameters which are not important for given task. The potential design of angular loss function is not unique, and further we discuss a few possible options for it.

## 2  Related Work

Methods of computational CC may be separated in two big groups: statistics-based and learning-based. Methods from the first group were widely used in last decades and exploit statistics of a single image. Moreover, in general they usually apply strong empirical assumptions, and operate in their limits. From these methods we can highlight the most important ones: Gray World [8] which is based on assumption that average color in the image is gray and tries to estimate color of illumination as shift causing non-gray average, White Patch [7] which is based on the assumption that the brightest point on image is a perfect white reflector and uses its color as color of illumination, Grey-Edge [38], and some more recent [10,40,21]. All of them were unified in a single framework by van de Weijer *et al.* [38].

Learning-based techniques estimate illumination color using a model created on a training dataset. In general, learning-based methods are shown to be more accurate than statistics-based approaches. This group includes Gamut Mapping algorithm [16], the svr-based algorithm [20], the exemplar-based algorithm [28], and numerous CNN-based algorithms presented in the last 3 years [4,35,30,19,27]. Many of neural network-based algorithms [11,9,15,34] use hand-crafted, low-level visual features, however most recent learn features using convolutional neural networks. Bianco *et al.* [4] firstly used patch-based CNNs for color constancy; in their work, simple CNN was used to extract local features which then were pooled [4] or passed to a support vector regressor [5]. Later, Shi *et al.* [35] proposed a more advanced network to deal with estimation ambiguities. Usage of the patches cropped from the images increases the size of the training dataset and augmentation of the data, however in cost of loss of semantic information. Algorithm of Lou *et al.* [30] works with full images and processes them with deep CNN that was pre-trained on a big ImageNet dataset with labels evaluated from hand-crafted color constancy algorithms and fine-tuned on each single dataset with ground truth labels. This work is the most relevant to the algorithm presented in this paper, but uses much simpler network as a base (AlexNet [29], while we use GoogLeNet[37]) and does not prove the necessity of the first step (in this paper higher accuracy was achieved without using any hand-crafted labels). In the work of Fourure *et al.* [19] custom Mixed Max-Minkowski pooling and Single Max Pooling networks were presented and demonstrated state-of-the-art accuracy. The latest and, to the best of our knowledge, most accurate algorithm was presented by Hu *et al.* [27] and is called $FC^4$. $FC^4$ is a fully-convolutional network that allows using images without resizing or cropping; also, it uses confidence-weighed pooling which helps to avoid ambiguity through assignment to each patch confidence weights according to the value they provide for color constancy estimation. Our algorithm, similarly to the algorithm of Lou *et al.* [30] is not fully-convolutional. This disadvantage, however, is not critical, because it can be solved in just one preprocessing step (resizing) with some loss of semantic information. Additionally, using any fully-convolutional net instead of GoogLeNet also solves this problem. In addition to the above, there are also a few more specifically oriented works, for instance, aimed on face regions [6], texture classification [3], or videos [33].

## 3 Experimental

### 3.1 Problem formulation

Following previous works, our goal was to estimate the color of illumination, noted as $e = (e_1, e_2, e_3)$, by given RGB image, to be able to discard illumination color cast using the von Kries [39] diagonal transform:

$$\begin{bmatrix} R_c \\ G_c \\ B_c \end{bmatrix} = \begin{bmatrix} e_1^{-1} & 0 & 0 \\ 0 & e_2^{-1} & 0 \\ 0 & 0 & e_3^{-1} \end{bmatrix} \begin{bmatrix} R \\ G \\ B \end{bmatrix}, \quad (1)$$

where $(R_c, G_c, B_c)$ is the corrected color as it appears under canonical white illumination. While there can be multiple illuminants in a scene, this work is focused on the traditional problem of estimating a single global illumination color, *e.g.* $e(x, y) = e$. Since we are not interested in the change in global intensity of illumination all the illumination vectors were normalized as follows:

$$e \leftarrow \sqrt{3} \frac{e}{\|e\|} \quad (2)$$

For comparison of predicted illumination vector ($\hat{e}$) and ground truth data ($e$) angular error metric (Eq. 3) is considered. This metric was first proposed by Hordley and Finlayson [25] and nowadays is a standard in this field.

$$\varepsilon = \arccos\left(\frac{\langle \hat{e}, e \rangle}{\|\hat{e}\|\|e\|}\right) \quad (3)$$

### 3.2 Network architecture

In the proposed approach GoogLeNet by Szegedy *et al.* [37] is used as a starting point for transfer learning. GoogLeNet is a 22 layers' deep network (Fig. 3) which achieved state-of-the-art accuracy in classification and detection in the ImageNet Large-Scale Visual Recognition Challenge 2014 (ILSVRC14). In comparison with AlexNet [29] it uses 12x less parameters, therefore works faster, and also provides higher accuracy. The core of this network is 9 Inception modules. The Inception module basically acts as multiple convolution filters, that are applied to the same input, with some pooling. The results are then concatenated. This allows the model to take advantage of multi-level feature extraction and to cover a bigger area, while keeping a fine resolution for small information on the images.

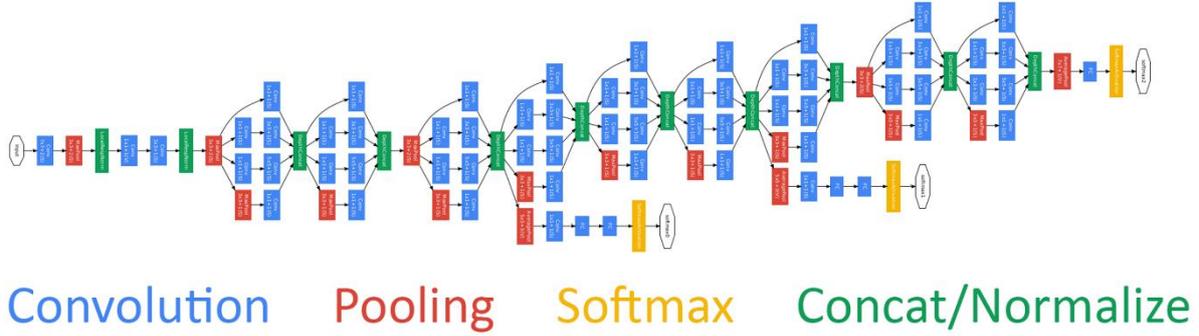

**Figure 2**. Schematic representation of the GoogLeNet. Credits [37].

GoogLeNet has been trained on over a million images and can classify images into 1000 object categories. To modify the network for regression task, first of all, we had to remove the last three layers, which contain information on how to combine the features that the network extracts into class probabilities and labels ("FC", "SoftmaxActivation", "Softmax2"). In their place, a fully connected layer with three neurons and a regression output layer were added.

### 3.2.1 Angular loss function

The MSE loss function is used in the regression layer by default. When using it, the model is trying to predict the illumination coordinates as close as possible to the ground truth data, *e.g.* predict the same vector. However, since lengths are normalized and only angles are important for CC task, we can benefit from it by removing restrictions on a length and adding a degree of freedom. Thus, by changing the loss function to the one which depends only on angle, we make the model more flexible and task oriented.

The primary value that the algorithm computes is a cosine of angle between predicted values and ground truth. Considering this fact, following loss functions were proposed:

$$L_1 = \arccos \cos \varepsilon = \varepsilon \tag{4}$$
$$L_2 = 1 - \cos \varepsilon \tag{5}$$
$$L_3 = 1 - \cos^2 \varepsilon = \sin^2 \varepsilon \tag{6}$$
$$L_4 = \sqrt{1 - \cos^2 \varepsilon} = \sin \varepsilon \tag{7}$$

Of course, choice is not limited to these options, but these are the simplest ones. To the best of our knowledge, only the $L_1$ function was used for color constancy by Hu *et al.* [27], and $L_2$ was used by Hara et al. [24] for a very different task. In the selection, we took into consideration the complexity of the function, its derivative, shape of the curve, and possible problematic points. Consequently, function $L_4$ was immediately discarded, due to the complexity, complexity of the derivative and the same behavior in the neighborhood of a 0 as $L_1$. Function $L_1$ is the most direct loss with respect to the error, and it also showed good result in case of FC[4] net. However, the fact that error is computed as arccosine of cosine makes the derivative much more complex (Eq. 8) and generates error or NaN values when $\varepsilon$ equals $0, \pi, 2\pi, \ldots$ (that was a major problem in our experiment).

$$\frac{\partial L_1}{\partial \hat{e}_i} = \frac{\partial(\arccos \cos \varepsilon)}{\partial \hat{e}_i} = -\frac{1}{\sqrt{1 - \cos^2 \varepsilon}} \cdot \frac{\partial \left(\frac{\langle \hat{e}, e \rangle}{\|\hat{e}\| \|e\|}\right)}{\partial \hat{e}_i} = \tag{8}$$

$$= \frac{1}{\sqrt{1-\frac{\langle \hat{e},e\rangle^2}{\|\hat{e}\|^2\|e\|^2}}} \cdot \frac{\hat{e}_i\langle \hat{e},e\rangle - e_i\langle e,e\rangle}{\|\hat{e}\|^3 \cdot \|e\|} = \frac{\hat{e}_i\langle \hat{e},e\rangle - e_i\langle e,e\rangle}{\langle \hat{e},\hat{e}\rangle\sqrt{\langle \hat{e},\hat{e}\rangle\langle e,e\rangle - \langle \hat{e},e\rangle^2}}$$

The expansion of the functions $L_2$ and $L_3$ in Maclaurin series (Eq. 9, 10) clearly demonstrates their behavior proportional to $\varepsilon^2$ around 0. We consider this feature beneficial for gradient computation by analogy with MSE loss.

$$1 - \cos\varepsilon = \frac{\varepsilon^2}{2} + \frac{\varepsilon^4}{24} - \frac{\varepsilon^6}{720} + O(\varepsilon^8) = 1 - \sum_{k=1}^{\infty} \frac{(-1)^k \varepsilon^{2k}}{(2k)!} \quad (9)$$

$$1 - \cos^2\varepsilon = \varepsilon^2 - \frac{\varepsilon^4}{3} + \frac{2\varepsilon^6}{45} - \frac{\varepsilon^8}{315} + O(\varepsilon^{10}) = -\sum_{k=1}^{\infty} \frac{(-1)^k (2^{-1+2k}\varepsilon^{2k})}{(2k)!} \quad (10)$$

Additional analysis of their plots (Fig. 4) reveals significant drawback of function $L_3$ – possibility to obtain an error of 180 degrees and negative values of illumination.

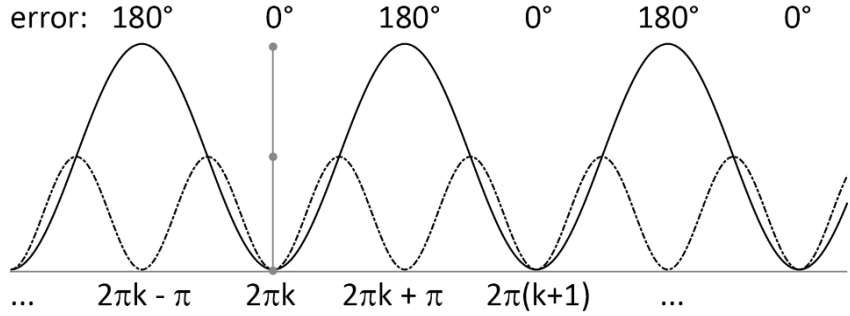

**Figure 3**. Plots of $y = 1 - \cos\varepsilon$ (solid line) and $y = 1 - \cos^2\varepsilon$ (dashed line)

Thereafter, loss function $L_2 = 1 - \cos\varepsilon$ was used in our experiments. A potential issue that derivative will become zero at point $\pi$ exists, but the probability of it is extremely low. Ultimately, derivatives of the functions $L_2$ and $L_3$ are following:

$$\frac{\partial L_2}{\partial \hat{e}_i} = \frac{\partial(1-\cos\varepsilon)}{\partial \hat{e}_i} = -\frac{\partial\left(\frac{\langle \hat{e},e\rangle}{\|\hat{e}\|\|e\|}\right)}{\partial \hat{e}_i} = \frac{\hat{e}_i\langle \hat{e},e\rangle - e_i\langle e,e\rangle}{\|\hat{e}\|^3 \cdot \|e\|} \quad (11)$$

$$\frac{\partial L_3}{\partial \hat{e}_i} = \frac{\partial(1-\cos^2\varepsilon)}{\partial \hat{e}_i} = -2\frac{\langle \hat{e},e\rangle}{\|\hat{e}\|\|e\|} \cdot \frac{\partial\left(\frac{\langle \hat{e},e\rangle}{\|\hat{e}\|\|e\|}\right)}{\partial \hat{e}_i} =$$
$$= 2\frac{\langle \hat{e},e\rangle}{\|\hat{e}\|\|e\|} \cdot \frac{\hat{e}_i\langle \hat{e},e\rangle - e_i\langle e,e\rangle}{\|\hat{e}\|^3 \cdot \|e\|} = 2\frac{\hat{e}_i\langle \hat{e},e\rangle^2 - e_i\langle e,e\rangle\langle \hat{e},e\rangle}{\langle \hat{e},\hat{e}\rangle^2 \cdot \langle e,e\rangle} \quad (12)$$

### 3.3 Image Datasets

Two standard benchmark datasets, the SFU Grayball [12] and the ColorChecker Reprocessed (other names: RAW dataset, 568-dataset, Gehler's dataset) [31,22] are used. The Grayball dataset contains 11346 real-world images. In each image, a gray ball is placed in the right-bottom of the image that allows to obtain the ground-truth illumination color. During training and testing, the gray ball has been removed from the image. The ColorChecker dataset contains 568 real-world images. The Macbeth ColorChecker chart is included in every scene acquired, thus ground truth illumination is known. Both in training and testing subsets ColorChecker chart have been removed.

Additionally, geometrical data augmentation was applied to both datasets. Namely, it consisted of random translations along X and Y axis up to 30 pixels, and horizontal reflections. The augmentation increase variance of training data and helps to prevent the network from overfitting and memorizing the exact details of the training images. Particularly for Color Constancy task, models greatly benefit from chromatic augmentation, e.g. casting the images with semi-random illumination, and changing corresponding ground-truth vectors. In this paper it was not used, but we imply that accuracy of presented algorithms may be increased using this technique [30,27,19].

The fact that proposed model contains fully connected layer imposes restriction on size of input image of 224 x 224 pixels. Hence, all the images were resized and central square areas were cropped.

### 3.4 Other experimental details

All the models have been implemented in MATLAB 2017b, using Neural Network Toolbox. The plainness and understandability of MATLAB code allows to create and directly use models like ours to wide audience, which we consider an undoubted advantage. The source code is openly accessible and can be downloaded by the following link: https://github.com/acecreamu/color-constancy-googlenet . Notwithstanding the fact that development of the algorithm and growth of its complexity have no limits, the simplest pure form of the given algorithm can fit in 30 lines of code.

The technological equipment used in the experiments consisted of only one laptop with Intel i7-7500U (2.7 GHz) CPU, 16 Gb RAM, and NVIDIA 950MX (2 Gb) GPU, which also supports the concept of the wide availability of presented methods.

## 4 Results

Following previous papers, 15-fold cross validation was used for Grayball [12] dataset. For much smaller ColorChecker [22] dataset only 3-fold cross validation was used in order to repeat conditions of previous experiments and compare the results objectively. Each time, corresponding dataset was partitioned into 15 or 3 subsets, and in a loop each of them was used as test set, after that the results of the all iterations were averaged. Such approach provides reliable evaluation of model's performance minimizing the influence of randomness.

Tables 1 and 2 present the comparison of our results with current state-of-the-art algorithms. Several standard metrics are reported in terms of angular error in degrees: mean, median, trimean or standard deviation, mean of the lowest 25% of errors, mean of the highest 25% of errors, and 95th percentile. For reasons unknown, very limited statistical data were reported in case of Grayball dataset, however there is no such problem in case of ColorChecker dataset.

**Table 1**. Results obtained on SFU Grayball dataset, and comparison with state-of-the-art methods. First two sections correspond to statistic-based and learning-based methods. Top 5 results are highlighted with shades of gray.

|  | Mean | Med | Std | Best 25% | Worst 25% | 95th perc |
|---|---|---|---|---|---|---|
| Gray World, Buchsbaum [8] | 7.9 | 7.0 | - | - | 15.2 | - |
| White Patch, Brainard and Wandell [7] | 6.8 | 5.3 | - | - | - | - |
| Shades-of-Gray, Finlayson and Trezzi [18] | 6.1 | 5.3 | - | - | - | - |
| 2nd-order Gray-Edge, van de Weijer et al. [38] | 6.1 | 4.3 | - | - | - | - |
| Gray Patches, Yang et al. [40] | 6.1 | 4.6 | - | 1.1 | 13.6 | - |

| | Mean | Med | Tri | Best 25% | Worst 25% | 95th perc |
|---|---|---|---|---|---|---|
| Local Surface Refl. Gao et al. [21] | 6.0 | 5.1 | - | - | 11.9 | - |
| 1st-order-Gray-Edge, van de Weijer et al. [38] | 5.9 | 4.7 | - | - | - | - |
| 3D Scene Geometry, Elfiky et al. [13] | 5.4 | 4.5 | - | - | - | - |
| Temporal sequence, Prinet et al. [32] | 5.4 | 4.6 | - | - | - | - |
| Single max pooling, Fourure et al. [19] | 5.2 | 4.5 | - | - | - | - |
| Mixed MaxL5 pooling, Fourure et al. [19] | 4.9 | 4.3 | - | - | - | - |
| Exemplar-Based, Joze et al. [28] | 4.4 | 3.3 | - | - | - | - |
| AlexNet Retrained, Lou et al. [30] | 3.9 | 3.0 | 3.3 | - | - | - |
| GoogLeNet + MSE | 2.55 | 1.91 | 2.19 | 0.58 | 5.59 | 6.78 |
| GoogLeNet + Angular Loss | 1.98 | 1.49 | 1.87 | 0.39 | 4.51 | 5.73 |

In case of both datasets, model with angular loss outperforms the one with MSE loss. The empirical comparison of the different angular loss functions and design on the new ones may be a subject for future research.

**Table 2**. Results obtained on reprocessed ColorChecker dataset, and comparison with state-of-the-art methods. First two sections correspond to statistic-based and learning-based methods. Top 5 results are highlighted with shades of gray.

| | Mean | Med | Tri | Best 25% | Worst 25% | 95th perc |
|---|---|---|---|---|---|---|
| White Patch, Brainard and Wandell [7] | 7.55 | 5.68 | 6.35 | 1.45 | 16.12 | - |
| Gray World, Buchsbaum [8] | 6.36 | 6.28 | 6.28 | 2.33 | 10.58 | 11.3 |
| 2nd-order Gray-Edge, van de Weijer et al. [38] | 5.13 | 4.44 | 4.62 | 2.11 | 9.26 | - |
| Shades-of-Gray, Finlayson and Trezzi [18] | 4.93 | 4.01 | 4.23 | 1.14 | 10.20 | 11.9 |
| Natural Image Statistics, Gijsenij et al. [23] | 4.19 | 3.13 | 3.45 | 1.00 | 9.22 | 11.7 |
| 19 Edge Moments, Finlayson et al. [17] | 2.80 | 2.00 | - | - | - | 6.90 |
| Edge-based Gamut, Barnard [1] | 6.52 | 5.04 | 5.43 | 1.90 | 13.58 | - |
| Pixel-based Gamut, Barnard [1] | 4.20 | 2.33 | 2.91 | 0.50 | 10.72 | 14.1 |
| Mixed MaxL5 pooling, Fourure et al. [19] | 3.33 | 2.63 | - | - | - | - |
| Single max pooling, Fourure et al. [19] | 3.31 | 2.59 | - | - | - | - |
| Exemplar-Based, Joze et al. [28] | 3.10 | 2.30 | - | - | - | - |
| AlexNet Retrained, Lou et al. [30] | 3.1 | 2.30 | - | - | - | - |
| Patch-CNN, Bianco et al. [4] | 2.63 | 1.98 | - | - | - | - |
| CCC, Barron [2] | 1.95 | 1.22 | 1.38 | 0.35 | 4.76 | 5.85 |
| DS-Net (HypNet+SelNet), Shi et al. [35] | 1.90 | 1.12 | 1.33 | 0.31 | 4.84 | 5.99 |
| SqueezeNet-FC4, Hu et al. [27] | 1.65 | 1.18 | 1.27 | 0.38 | 3.78 | 4.73 |
| GoogLeNet + MSE | 3.06 | 2.15 | 2.32 | 0.48 | 7.01 | 9.64 |
| GoogLeNet + Angular Loss | 2.55 | 1.69 | 1.92 | 0.39 | 6.03 | 6.70 |

The visual evaluation of results can be done using the samples illustrated on Figure 4.

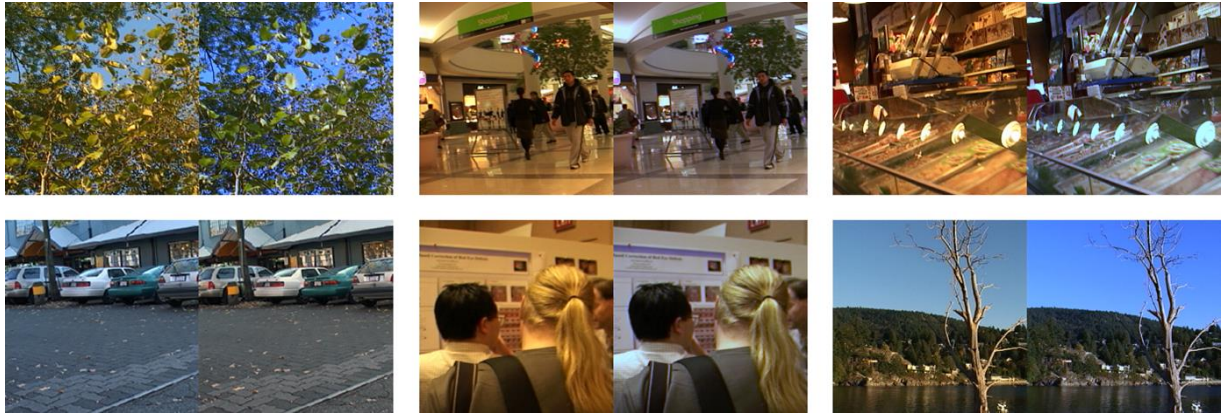

**Figure 4**. An example of images from Grayball dataset before (left) and after (right) removing illumination color cast using algorithm presented in this paper.

## 5 Conclusion

In this paper, we proposed an approach which allows to easily create effective Color Constancy algorithm. Presented technique may be applied to any advanced CNN. In our case we used GoogLeNet, and depending on the dataset obtained the best or comparable with the best accuracy. A significant advantage of our approach is that it requires neither high skills in machine learning, nor expensive technical equipment, nor a long time, which makes it available to a general public.

Also, we discussed design of angular loss function, which is an important question for any CC algorithm. In result, we chose function $1 - \cos \varepsilon$, where $\varepsilon$ is an angular error, because of its simplicity, efficiency, and high suitability for this task. However, the discussion and empirical examination are not finished, and may be extended in future works.